\documentclass[letterpaper, 10 pt, conference]{ieeeconf}  

\IEEEoverridecommandlockouts                              

\usepackage[hidelinks]{hyperref}
\usepackage{graphicx}
\usepackage{amsmath}
\usepackage{amssymb}
\usepackage{bm}
\usepackage{xcolor}
\usepackage{algorithm}
\usepackage{calc}
\usepackage{algpseudocode}
\newcommand{\Sset}{\mathcal{S}}
\newcommand{\Aset}{\mathcal{A}}

\newcommand{\state}{\mathbf{s}}

\newcommand{\svec}{\mathbf{s}}
\newcommand{\action}{\mathbf{a}}

\newcommand{\de}{\mathrm{d}}

\newcommand{\omegavec}{\bm{\omega}}
\newcommand{\Omegavec}{\bm{\Omega}}
\newcommand{\muvec}{\bm{\mu}}

\newcommand{\phivec}{\bm{\phi}}

\newcommand{\epsilonvec}{\bm{\epsilon}}

\newcommand{\aligntoright}[1]{
\settowidth{\eql}{$#1$}
\setlength{\eql}{\linewidth-\eql}
\hspace*{\eql}\ignorespaces 
#1}

\title{\LARGE \bf
Dynamic Decision Frequency with Continuous Options
}

\author{Amirmohammad Karimi$^{1, \dagger}$, Jun Jin$^{3}$, Jun Luo$^{3}$, A. Rupam Mahmood$^{1,4}$, Martin Jagersand$^{1}$, Samuele Tosatto$^{2}$
\thanks{$^{1}$ Department of Computer Science, University of Alberta, Edmonton, Canada.%
$^{2}$ Department of Computer Science and Digital Science Center, University of Innsbruck, Innsbruck, Austria.}%
\thanks{$^{3}$Noah's Ark Lab, Huawei Technologies Canada, Edmonton, Canada.}
\thanks{$^{4}$Alberta Machine Intelligence Institute (Amii)}
\thanks{$^\dagger$ Corresponding author: \tt\small amirmoha@ualberta.ca}}

\usepackage{pgfplots}
\usepackage{tikz}
\usepackage{subfig}
\usetikzlibrary{positioning}\usepgfplotslibrary{groupplots}
\usetikzlibrary{shadows}
\usetikzlibrary{shapes.multipart}
\usetikzlibrary{fit}
\usetikzlibrary{decorations.pathmorphing} 
\usetikzlibrary{backgrounds} 
\newtheorem{theorem}{Theorem}
\begin{document}

\maketitle
\thispagestyle{empty}
\pagestyle{empty}

\begin{abstract}
In classic reinforcement learning algorithms, agents make decisions at discrete and fixed time intervals. The duration between decisions becomes a crucial hyperparameter, as setting it too short may increase the problem’s difficulty by requiring the agent to make numerous decisions to achieve its goal while setting it too long can result in the agent losing control over the system. However, physical systems do not necessarily require a constant control frequency, and for learning agents, it is often preferable to operate with a low frequency when possible and a high frequency when necessary. We propose a framework called Continuous-Time Continuous-Options (CTCO), where the agent chooses options as sub-policies of variable durations. These options are time-continuous and can interact with the system at any desired frequency providing a smooth change of actions. We demonstrate the effectiveness of CTCO by comparing its performance to classical RL and temporal-abstraction RL methods on simulated continuous control tasks with various action-cycle times. We show that our algorithm's performance is not affected by the choice of environment interaction frequency. Furthermore, we demonstrate the efficacy of CTCO in facilitating exploration in a real-world visual reaching task for a 7 DOF robotic arm with sparse rewards.
\end{abstract}

\section{Introduction}

Reinforcement Learning (RL) has become an effective strategy for solving complex control tasks, including continuous control and robotic manipulation, as shown by its success in recent years \cite{sutton_reinforcement_2018,lillicrap_continuous_2016,haarnoja_soft_2018,kober_reinforcement_2011}. Although these tasks are typically formulated using discrete-time Markov Decision Processes (MDPs) \cite{puterman2014markov}, which assumes that control signals are issued at specific time intervals, real-world problems are often better defined in the continuous-time domain. Despite some studies on RL in continuous-time MDPs, the majority of research has been concentrated on the discrete-time formulation, which serves as a useful approximation but may not be optimal for all problems \cite{yildiz2021continuous,xiao2019thinking}.

RL agents typically select actions at discrete time intervals. This approach has some drawbacks. As the time interleaving two consecutive actions approaches zero, the effect of each action on the environment becomes negligible, making it difficult for the agent to determine the optimal action \cite{baird1994reinforcement,tallec2019making}.  
Furthermore, high-frequency decision-making increases the effective time horizon of the problem (i.e., more decisions are needed to reach the goal state) resulting in high sample complexity and challenging exploration, especially in goal-based tasks or sparse rewards settings \cite{amin2020locally,park_time_2021}. In contrast, low-frequency decision-making often exhibits better exploration but hinders the controllability of the environment. These challenges raise the question of what the agent's decision frequency should be. In this paper, we argue that the decision frequency does not necessarily need to be fixed. Typically, the decision frequency should be higher when the system is hard to control, while it can be lower otherwise~\cite{kleff2021high}. A good example is the swing-up pendulum task. During the ``swinging phase'', the system is simple to control, and one can decide to swing left or right with low frequency until the pendulum is upright. In the stabilization phase, however, the system is unstable, and one needs to apply frequent small torques in different directions to compensate for perturbations. 
In other words, the learning algorithm can be more efficient by allowing the agent to make decisions less frequently when possible and more frequently when necessary.

In RL, control tasks are typically represented as MDPs with fixed action duration. The concept of explicitly choosing the duration of extended actions has been studied in a body of works for adapting the control frequency, which is known as action repetition RL (ARRL) methods \cite{lakshminarayanan_dynamic_2017, metelli2020control, sharma_learning_2017}. In ARRL, the action space is extended with a discrete variable that indicates the number of steps with which an action must be repeated. ARRL provides a na\"{i}ve temporal abstraction that increases sample efficiency. However, ARRL comes with some limitations. The discrete formulation of the action length does not scale in the high-frequency domain, where the action must be extended with a considerable number of discrete time steps to allow temporal abstraction. As we will show in the empirical analysis, ARRL does not perform well when the control frequency is high. The critical concept we develop in this paper is that time-continuous systems should be modeled with a continuous-time duration of actions. Furthermore, policies based on repeating an action are limited in representational power and exploration, especially in continuous control tasks where a smoothly changing control signal may be desirable. In real-world tasks, we often want to change the controller continuously to allow less aggressive and more energy-efficient controllers. As an example, recall the swing-up pendulum experiment: the pendulum could be swung upwards by applying the same action repeatedly, but a more energy-efficient solution would be to decrease the applied torque as the pendulum approaches the goal position. 

Hierarchical RL (HRL) provides a more sophisticated abstraction mechanism, where a high-level policy selects \textsl{options} containing lower-level policies, that elapse for multiple numbers of time-steps \cite{sutton_between_1999,bacon_option-critic_2017,zhang_dac_2019}. HRL exhibits better exploration compared to ARRL and enables gradual changes in the action while still allowing temporal abstraction. HRL, however, is typically hard to train due to its complex formulation and redundancy (as the optimal policy can always be expressed by one option, making the usage of more options redundant) \cite{harb2018waiting}. Furthermore, HRL is still inherently discrete-time and exhibits performance degradation when applied to high-frequency control settings.

To overcome the aforementioned issues, in this paper, we propose Continous-Time Continuous-Options (CTCO), a novel temporal abstraction mechanism based on a continuous-time architecture for reinforcement learning that allows a continuous, smooth change of action.  
Our algorithm recalls the options framework introduced by Sutton et al. in 1999 \cite{sutton_between_1999} as it uses optionfs; however, in our work, such options have continuous-time open-loop policies that are applied to the environment for a duration decided by the agent's policy. Open-loop policies avoid the redundancy problem of classic HRL. Although a single open-loop controller may not solve the task, the agent's policy can switch between different open-loop controllers when needed. The continuous-time duration of our controllers contrasts with ARRL methods, which select action duration from a discrete, finite set. The choice of continuous option durations is not sensitive to the environment's control frequency making the algorithm robust w.r.t. this hyper-parameter. 

To summarize, CTCO is defined in continuous time to allow scalability across different interaction frequencies. It allows a temporal mechanism to be seen as a hybrid between HRL and ARRL. Similarly to HRL, our algorithm is based on options, but unlike the classic options framework (and similarly to ARRL), such options have open-loop policies, decreasing the system redundancy. Our implementation of CTCO is based on the soft actor-critic (SAC) \cite{haarnoja_soft_2018}.
We analyze the robustness of our framework w.r.t. different interaction frequencies and compare it against classical, hierarchical, and action repetition RL methods. We find that, while the considered baselines exhibit performance degradation when tested with high interaction frequencies, CTCO is not subject to any performance degradation. Furthermore, we test CTCO on a real-world robotic task without any sim-to-real pertaining, showing that the higher sample efficiency and the use of smooth open-loop controllers in CTCO are particularly well suited for real-world robotics.
\section{Background}\label{sec:background}
\textbf{Discrete-Time Decision-Making} is formalized using Markov decision processes (MDPs) $\langle \Sset, \Aset, p, r, \gamma, \mu_0 \rangle$, where $\Sset$ is the set of states; $\Aset$ is the set of actions; $p(\state’| \state, \action)$ is the probability of visiting $\state’ \in \Sset$ after the application of action $\action \in \Aset$ to state $\state \in \Sset$; $r(\state, \action)$ is the reward signal in state $\state$ and action $\action$, $\gamma$ is the discount factor and $\mu_0$ is the distribution  of starting states. \textsl{Decisions} are often encoded with a stochastic parametric model $\action \sim \pi_\theta(\state)$ where $\theta$ is the set of parameters. The goal of reinforcement learning is to find the set of parameters that maximizes the discounted objective
\begin{align}
J^{\gamma}_\pi(\theta) := \mathbb{E}_\pi\left[\sum_{t=0}^{\infty}\gamma^t r(\state_t, \action_t)\right]. \nonumber
\end{align}
In discrete time decision-making, there is no notion of physical time: the time elapsed between two consecutive state observations is not considered. However, when implemented on the real system, the designer needs to decide \textsl{when} the system needs to observe the environment and make decisions. Since the physical time is not considered in this mathematical mode,  the time interleaving two observations (or decisions) is usually kept constant. 
The time interleaving two consecutive decisions determines the learning agent's frequency, and it heavily impacts the performances \cite{mahmood_setting_2018}.

\textbf{Continuous-Time Decision-Making} assumes a continuous evolution of states, actions, and rewards defined with a \textsl{time-independent} dynamical system~\cite{doya2000reinforcement}
\begin{align}
   \dot{\state_t} = f(\state_t, \action_t) \label{eq:dynamical-system}.
\end{align}
For simplicity, Equation~\ref{eq:dynamical-system} does not consider stochasticity, which can be formalized with a more rigorous set of assumptions. Informally, such assumptions ensure that the probability of observing a state $\state_t$ after a continuous sequence of actions $\action_{0:t}$ starting from a given state $\state_t$ is Markovian, i.e., 
\begin{align}
	p(\state_{t+d} | \state_{t}, \action_{t:t+d}) = p(\state_{t+d} | \state_t, \action_{0:t+d}). \nonumber 
\end{align}
where $d>0$ denotes the duration of the sequence.
Furthermore, the system is time-invariant, i.e., 
\begin{align}
	p(\state_{t_1+d} | \state_{t_1}, \action_{t_1:t_1+d}) = p(\state_{t_2+d} | \state_{t_2}, \action_{t_2:t_2+d})\nonumber\\  \forall t_1, t_2 \geq 0 \ s.t \ \state_{t1} = \state_{t2}, \action_{t_1:t_1+d}=\action_{t_2:t_2+d}. \nonumber 
\end{align}
In infinite-horizon problems, the discounted objective becomes
\begin{align}
J^\tau_\pi(\action_{0:\infty}) = \mathbb{E}\left[\int_0^\infty e^{-\tau t} r(\state_t, \action_t)\de t\right]. \label{eq:continuous-objective}
\end{align}
\textbf{A Unified View.} The continuous system described can be viewed as a discrete system too \cite{park_time_2021}. This second view is more convenient because it allows using existing tools for discrete processes to optimize the continuous systems. Consider the following modified MDPs, $\langle\Sset, \Aset^\infty, P, R, \tau, \mu_0 \rangle$, where $\Aset^\infty$ is the set of continuous action sequences, (informally, $\Aset^\infty \equiv \mathcal{P}(\{\action_{t:t+\Delta t} : \forall t , \Delta t > 0\})$ where $\mathcal{P}(x)$ is the \textsl{power set} of $x$), and $\tau > 0$ a time-constant. In this system, the reward becomes
\begin{align}
R(\state_t, \action_{t:t+d}) = \int_{t}^{t+d} e^{-\tau (\kappa -t)} r(\state_\kappa, \action_\kappa)\de \kappa, \nonumber 
\end{align}
the transition becomes $P(\state_{t+d} | \state_t, \action_{t:t+d})$ where $\state_{t+d} = \state_t + \int_{t}^{t+d}\dot{\state}_x \de x$, and the discount factor becomes variable, i.e., 
\begin{align}
\gamma(d) = e^{-\tau d}. \nonumber 
\end{align}
By choosing a set of discretization points $t_1 < t_2 < t_3, \dots$, the objective function from \ref{eq:continuous-objective} can be rewritten as
\begin{align}
J^\tau_\pi(\theta) = \mathbb{E}_{\pi}\left[\sum_{i=0}^\infty\left(\prod_{j=0}^{i}\gamma(d_j)\right) R(\state_{t_i}, \action_{t_i:t_{i+1}})\right], \label{eq:unified-objective} 
\end{align}
where $d_i = t_{i+1} - t_i$.
Notice that Equation~\ref{eq:unified-objective} is equivalent to Equation~\ref{eq:continuous-objective}. This discretization is well known, e.g., \cite{park_time_2021}. It is interesting to notice that MDPs with variable discounting have been introduced in \cite{white_unifying_2017} where the author aims to provide a unified framework for RL. However, in White’s work, the variable discount factor is treated as part of the environment, and its effect is studied only on the value estimation. In this work, as we will see, the discount factor is, in practice, decided by the high-level policy (which is devoted to choosing low-policy durations). We also study its effect on the policy gradient update rule.

\section{Method}\label{sec:method}
Our framework, Continuous-Time Continuous-Options (CTCO), comprises a policy $\pi$, and a set of options encoded as a parametric model $\Omegavec(\omegavec)$. The policy $\pi$ selects options with variable durations and decides which option to execute only after the termination of the current option, allowing the system to dynamically determine the decision frequency. However, different option durations may have the same effect on the environment (e.g., an option that outputs action $\action=1$ for $3$ seconds is equivalent to three options that output action $\action=1$ for one second each). Since longer durations are preferable for the learner in case of ambiguity ( it requires less number of \textsl{decisions}), we introduced a regularization factor that favors longer durations. We implemented the update rule of our framework by taking inspiration from soft actor-critic (SAC) algorithm, a state-of-the-art policy gradient method that includes entropy regularization to achieve efficient exploration \cite{haarnoja_soft_2018}.
\subsection{The Options}
In Section \ref{sec:background}, we saw that continuous time (time-independent) dynamical systems can be seen as particular MDPs where actions are replaced with continuous sequences of actions (i.e, $\action_{t:t+d}$), and the discount factor becomes variable. 
However, it is impossible to represent continuous sequences with a finite memory. One way to mitigate this issue is to parameterize these sequences as
\begin{align}
	\action_{t} = \Omegavec(\state_t, \omegavec, t, d) : t \in [0, d] , \nonumber
\end{align}
where the option $\Omegavec$ can be implemented with any formula, neural network, or program that depends on a vector of parameters $\omegavec$, the current state $\state_t$, time $t$, and duration variable $d$. Motivated to obtain a simple model with few parameters that can be computed fast (which is ideal for real-time computations), we encoded options using a linear parametric model without state dependency 
\begin{align}
	\action_{t} = \Omegavec(\omegavec, t, d) = \phivec^\intercal(t/d)\omegavec : t \in [0, d]. \label{eq:option}
\end{align}
Inspired by work on movement primitives, we opt to encode the features $\phivec$ with normalized radial basis functions (RBFs), ensuring that the sequence of actions is \textsl{smooth} and produces movements that contain low jerkiness and are ideal for robotic applications. The number of RBFs determines the complexity of the movement. With one RBF, we obtain a constant action output (similar to ARRL), while with more RBFs, we obtain a more complicated output. We designed option policies without the dependency on the state $\state_t$ opting for open-loop controllers. However, the policy $\pi$, compensates for the lack of feedback since it selects options by observing the current state of the environment $\state_t$.
\subsection{The Policy}
The policy $\pi$ is responsible for choosing the parameters $\omegavec$ and the duration $d$ of the options. In particular, the policy receives as input the current state of the systems $\state$ and determines the probability density of the parameter vector and the duration independently, i.e., 
\begin{align}
	\omegavec, d \sim \pi_{\theta}(.,.|\state).\nonumber
\end{align}
The probability density function of $\omegavec$ is modeled as a multidimensional Gaussian distribution while the density function of $d$ is a transformation of Gaussian distribution to a pdf over positive numbers by applying an invertible function; we choose the sigmoid function. 
We parameterize the policy using a neural network that outputs $\muvec^{\omegavec}_{\theta}(\state_t)$, $\boldsymbol{\sigma}^{\omegavec}_{\theta}(\state_t)$ , $\mu^{d}_{\theta}(\state_t)$, $\sigma^{d}_{\theta}(\state_t)$ as mean and variance of the Gaussian distributions to sample the parameters vector  $\omegavec$ and the duration $d$ given observation of state $\state_t$. To sample  $\omegavec$ and $d$, and compute the gradients of the objective (eq. \ref{eq:unified-objective}) w.r.t. the policy parameters we use the reparametrization trick, %
\begin{align}
    &\omegavec = \bm{f}^{\omegavec}_\theta(\state; \bm{\epsilon}) := \muvec^{\omegavec}_{\theta}(\state) + \bm{\epsilon} \bm{\sigma}^{\omegavec}_{\theta}(\state), \nonumber\\
    d = f&^{d}_\theta(\state; \epsilon) := d_{max}\mathrm{sigm}(\muvec^{d}_\theta(\state) + \epsilon \bm{\sigma}^{d}_\theta(\state)), \nonumber
\end{align}
with $\epsilon,\bm{\epsilon} \sim \mathcal{N}(\mathbf{0}, \mathbf{I})$. Here $d$ is limited to continuous values in $(0, d_{max})$. In what follows the evaluation and improvement of the parameterized policy is described. 

\begin{algorithm}[t]
\caption{Continuous-Time Continuous-Option}\label{alg:ctco}
\smallskip
\begin{algorithmic}
\Require a policy $\pi$ with a set of parameters $\theta, \theta'$, critic parameters $\chi, \chi'$, option model $\Omega$, learning-rates $\lambda_q, \lambda_p$, replay buffer $\mathcal{B}$
\State $i = 0$, $t_i =  0$, observe $\state_0$
\While{True}
\State $\omegavec_i, d_i \sim \pi_{\theta}(\state_{t_i})$ \Comment{sample option and duration}
\State Execute $\Omegavec( \omegavec_i, \kappa ,d_i)$ for $ \kappa = 0 \text{ to } d_i$ seconds 
\State $i \gets i + 1$ 
\State $t_i = t_{i-1} + d_{i-1}$
\State Observe $\state_{t_i} $ and compute $R_{i-1}$ with \eqref{eq:continuous-reward}
\State Store $\state_{t_{i-1}}, \omegavec_{i-1}, d_{i-1}, R_{i-1}, \state_{t_i}$ in $\mathcal{B}$
\State Sample $\state, \omegavec, d, R, \state'$ from $\mathcal{B}$
\State $\chi \gets \chi - \lambda_q \nabla_\chi \mathcal{L}_{Q}(\chi, \state, \omegavec, d, R, \state')$ \Comment{critic update}
\State $\theta \gets \theta - \lambda_p \nabla_\theta \mathcal{L}_{\pi}(\theta, \state, \omegavec, d)$ \Comment{actor update}
\State Perform soft-update of $\chi'$ and $\theta'$
\EndWhile
\end{algorithmic}
\end{algorithm}
\subsection{A Soft Actor-Critic Framework}
From the perspective of the discretized MDP introduced in Section~\ref{sec:background}, $(\omegavec , d)$ represents the continuous sequence of actions $\action_{t:t+d}$, therefore, can be considered as the action in the discretized MDP. Taking into account this consideration, the reward definition in Section~\ref{sec:background} and the option policy definition in \eqref{eq:option}, we obtain the reward that depends on state, option parameters, and option duration, 
\begin{align}
	R(\state_t, \omegavec, d) &= \int_{t}^{t+d} e^{-\tau  (\kappa -t)} r(\state_\kappa, \Omegavec(\omegavec, \kappa, d))\de \kappa. \label{eq:continuous-reward}
\end{align}
Notice that $R(\state, \omegavec, d)$ cannot, in general, be computed in closed-form, but it can be approximated via numerical integration.
In this MDP, the overall objective becomes
\begin{align}
	J^\tau_\pi(\theta) = \mathbb{E}_{\pi}\left[\sum_{i=0}^\infty\prod_{j=0}^{i-1}\gamma(d_j) R_i\right], \nonumber
\end{align}
where $R_i=R(\state_{t_i}, \omegavec_{i}, d_{i})$, $t_i = \sum_0^{i}d_i$, $d_i$ are option durations chosen by the policy $\pi$, and $\state_{t_i}$ are the observed states. 

Until now, we have described a mathematical framework that shows how a policy $\pi_\theta$ interacts with a continuous environment. Now we define how such policy can be improved. Among many different options, we choose to implement our algorithm following the soft actor-critic (SAC) architecture \cite{haarnoja_soft_2018}.
To this end, we include an entropic regularization term that encourages the exploration of different option parameters $\omegavec$ and durations $d$. Furthermore, we include a new component that penalizes short options. This additional regularizer, called \textsl{high-frequency penalization} is needed, since the learner can find solutions that work well with arbirtarly high frequency, but such high frequency is detrimental for training performances \cite{harutyunyan2019termination}. The high-frequency penalization consists of a constant term subtracted from the objective each time the policy makes a decision. The overall objective incorporates both the entropic and the frequency regularization terms,
\begin{align}
	\!J^\tau_\pi(\theta) \! =\! \underset{\pi}{\mathbb{E}}\! \left[\sum_{i=0}^\infty\prod_{j=0}^{i-1}\! \gamma(d_j)\left( R_i\! +\!\beta_E \mathcal{H}(\pi_\theta(\cdot , \cdot \mid \!\state_{t_i}))\! -\! \beta_h\right)\! \right]\!\! \nonumber
\end{align}
where $\beta_E$ and $\beta_h$ are respectively the entropic and the high-frequency regularizers. In addition to these regularization terms, following the SAC architecture, we introduce an approximator for the $Q$-function $\hat{Q}_{\chi}$, a target $Q$-function $\hat{Q}_{\chi'}$ and a target policy $\pi_{\theta'}$.

\textbf{Policy Evaluation.}
The $Q$-function evaluates the policy at any state-action pair. In CTCO, the $Q$-function is described by the following Bellman equation,
\newlength{\eql}
\begin{align}
    &Q^\pi(\state, \omegavec, d) =\underset{\omegavec',d',\state'}{\mathbb{E}}\bigg[ R(\state, \omegavec,d) - \beta_h + \gamma(d) (Q^\pi(\state', \omegavec', d')\nonumber
    \\&\aligntoright{ -\beta_E \log \pi_\theta(\omegavec', d' | \state'))\bigg]}\label{eq:bellman}
\end{align}    
where $\omegavec', d' \sim \pi_{\theta'}(\state')$ and $\pi_{\theta'}$ is the target policy. 
Equation~\ref{eq:bellman} cannot be solved in closed form. Popular methods based on temporal difference introduce two function approximators $\hat{Q}_\chi$ and $\hat{Q}_{\chi'}$ that tabilize learning $Q$-function \cite{ernst_tree-based_2005,mnih_human-level_2015}. Similarly to SAC, our algorithm minimizes the mean square Bellman error
\begin{align*}
    & \mathcal{L}_{Q}(\chi, \state, \omegavec, d, R, \state') = \Big(R \nonumber -\beta_h 
    + \gamma(d)(\hat{Q}_{\chi'}(\state', \omegavec', d')\\
    &\aligntoright{-\beta_E \log \pi(\omegavec', d' | \state')) -\hat{Q}_\chi(\state, \omegavec, d)\Big)^2}
\end{align*}    
and use a soft-update rule to update the parameters $\chi$ and $\chi'$ (i.e., $\chi' = (1-\alpha_\chi) \chi' + \alpha_\chi \chi$).

\textbf{Policy improvement.}
The classic policy gradient theorem does not consider situations where the policy changes the action duration and the discount factor. Therefore we derived the policy gradient with this new assumption,
\begin{align}
    &\nabla_{\theta} J^\tau_\pi(\theta)
    \propto \underset{\mu_\pi}{\mathbb{E}}\bigg[\nabla_{\omegavec} Q^\pi(\state, \omegavec, d)\nabla_\theta \bm{f}^{\omegavec}_\theta(\svec, \epsilonvec) + \! \nabla_d Q^\pi(\state, \omegavec, d)\nonumber\\
    &\aligntoright{\nabla_\theta \bm{f}^{d}_\theta(\svec, \epsilonvec) 
     - \beta_E \nabla_{\theta} \log \pi_\theta(\omegavec, d| \state)\Big|_{\omegavec =\bm{f}^{\omegavec}_\theta	(\svec, \epsilonvec), d = {f^{d}_\theta(\svec, \epsilon)}}\bigg].}\label{eq:policy-gradient}
\end{align}
The proof of policy gradient derivation can be found in \ref{sec:appendix}. The policy gradient in Equation~\ref{eq:policy-gradient} allows estimating the gradient from samples, as in classic actor-critic frameworks. Hence, one can minimize the following surrogate objective
\begin{align}
    &\mathcal{L}_\pi(\theta, \state, \omegavec, d) =  \nabla_{\omegavec} Q^\pi(\state, \omegavec, d)\big|_{\omegavec =\bm{f}^{\omegavec}_\theta	(\svec, \epsilonvec)}\nabla_{\theta}\bm{f}^{\omegavec}_\theta(\svec, \epsilonvec)+ \nonumber  \\
	 &\aligntoright{\nabla_d Q^\pi(\state, \omegavec, d)\big|_{d = {\bm{f}^{d}(\svec, \epsilon)}
	}\nabla_{\theta}\bm{f}^{d}_\theta(\svec, \epsilonvec)- \beta_E \nabla_{\theta} \log \pi_\theta(\omegavec, d | \state).} \label{eq:actor-loss}
\end{align}
A scheme of CTCO is presented in Algorithm~\ref{alg:ctco}.
\section{Empirical Analysis}

\begin{figure*}[t!]
\centering
    \includegraphics[width=\textwidth]{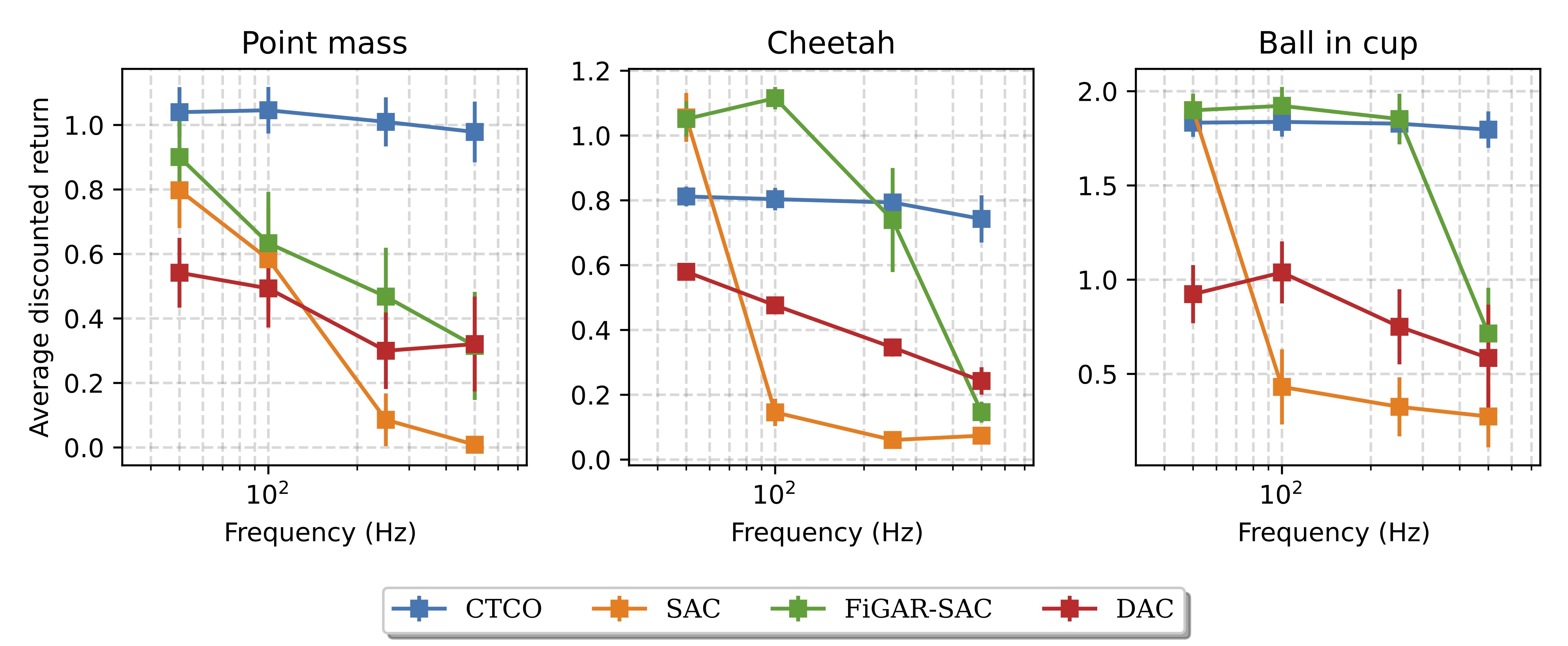}
\centering
\caption{\textbf{Frequency analysis plots}. CTCO's performance is not dampened by high interaction frequency with the environment. Bars depict 95\% confidence intervals. \label{fig:frequency-analysis}}
\end{figure*}

In this section, we focus on answering two questions:
\begin{enumerate}
    \item Is our algorithm robust w.r.t. high interaction frequency with the environment?
    \item How does our algorithm perform on a real-robotic task?
\end{enumerate}

\textbf{Comparisons.}
We compare our algorithm against classic RL by using soft actor-critic (SAC), ARRL by using fine-grained action repetition (FiGAR-SAC) \cite{sharma_learning_2017}, and HRL by using double actor-critic (DAC-PPO) \cite{zhang_dac_2019}. The algorithms are modified to simulate an asynchronous interaction with the environment.

\textbf{Experimental Setup.}
When interacting with the real world, the amount of computation that a learning algorithm performs is independent of the task execution. While in the simulation, we can perform a fixed number of updates between two different \texttt{step()} calls on the simulator, in the real-world, action signals and algorithm updates are asynchronous  \cite{yuan2022asynchronous}. In this section, we study how different interaction frequencies impact the algorithms. To emulate a realistic effect of different interaction frequencies, we modified the simulated environments accordingly, and we modified the algorithms to keep a fixed number of updates per time unit. 
Furthermore, the discount factor of the discrete-time baselines (SAC, FiGAR-SAC and DAC-PPO) needs to be revised to keep a consistent performance metric across different frequencies. This scaling is obtained by setting $\gamma = \exp-\tau dt$ where $\tau = -\log \gamma_\text{base}/dt_\text{base}$ and $\gamma_\text{base}=0.98$ is the chosen discount factor for a fixed time interval $dt_\text{base}=0.05s$. The algorithms are implemented using \texttt{PyTorch} and \texttt{NumPy}. The structure of the actor is kept identical across the different baselines, except for the last layer, which needs to be adapted to the correct output size for our algorithm. In detail, the actor neural network has two hidden layers of 10 neurons. The critic neural network, instead, has two hidden layers of 64 neurons. The implementations are publicly available\footnote{\href{https://github.com/amir-karimi96/continuous-time-continuous-option-policy-gradient}{https://github.com/amir-karimi96/continuous-time-continuous-option-policy-gradient}}.
\subsection{Robustness to interaction frequencies}
In classic RL, interaction and decision frequencies are the same. Due to this design choice, high interaction frequencies negatively impact performance. 
In this experiment, we test SAC, FiGAR-SAC, DAC-PPO and our algorithm for different interaction frequencies on three different environments: \texttt{Point Mass}, \texttt{Cheetah} and \texttt{Ball in a Cup} from DeepMind control suit \cite{tassa2018deepmind}.
We hypothesize that in continuous control, RL frameworks which choose an action for each task time-step are not robust to high-frequency interactions. One issue arises in terms of exploration that is when the action-cycle time becomes small, the change in the state vanishes and the agent cannot explore the task state-space efficiently, given that the behaviour of the actor in the first stages of learning is independently random in each time-step. To examine this hypothesis, we set up the tasks of \texttt{Point Mass} and \texttt{Ball in a Cup} with sparse rewards, and \texttt{Cheetah} with dense rewards for different action-cycle times and measure discounted returns. For each algorithm, we run 30 seeds for 400 minutes of task time. For CTCO we use high-frequency penalty $\beta_h=0.05$ and options with $n_\text{RBF}=3$ in all tasks. Fig.~\ref{fig:frequency-analysis}a shows that the performance of SAC, FiGAR-SAC and DAC-PPO is influenced by interaction frequency. However, Our algorithm maintains almost constant performance across the range of different frequencies. In \texttt{Cheetah}, it has sub-optimal performance, suggesting that with dense reward, simpler algorithms like SAC and FiGAR-SAC achieve higher performance. 

\subsection{Real-World performance analysis}
\begin{figure}
\centering
    \subfloat[\label{fig:franka_task_setup}]{\includegraphics[height=6.5cm]{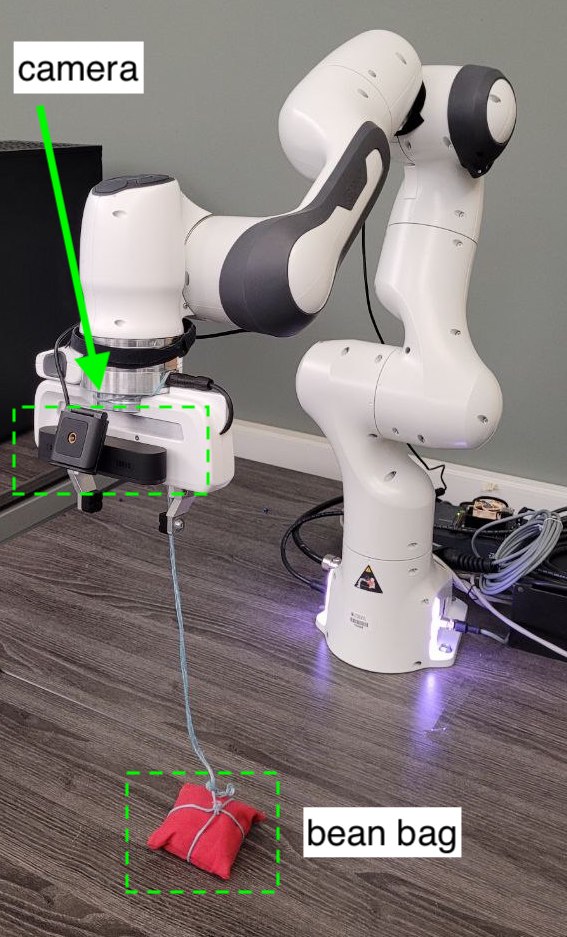}
    \includegraphics[height=6.5cm]{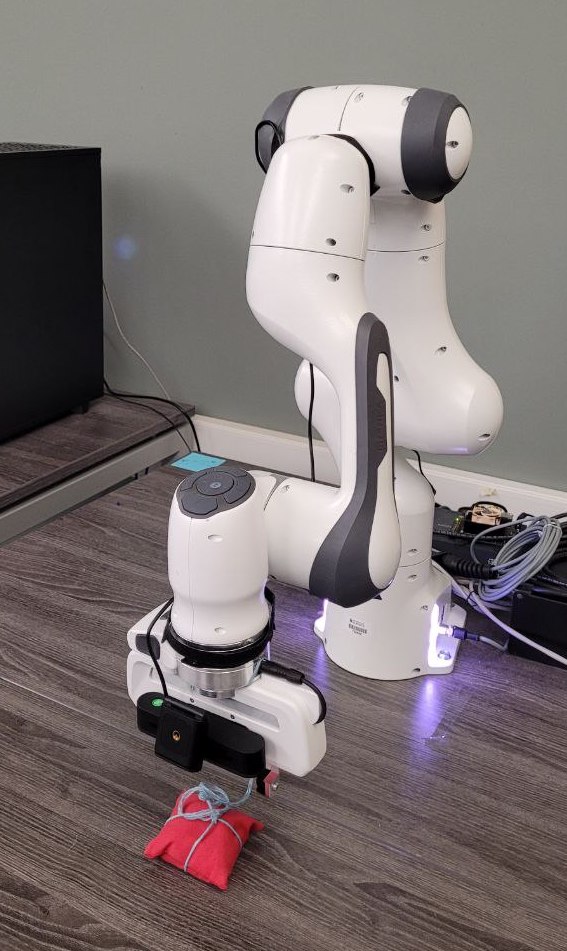}}
    \hfill
    \subfloat[\label{fig:franka_task_result_2}]{
    \includegraphics[width=0.5\textwidth]{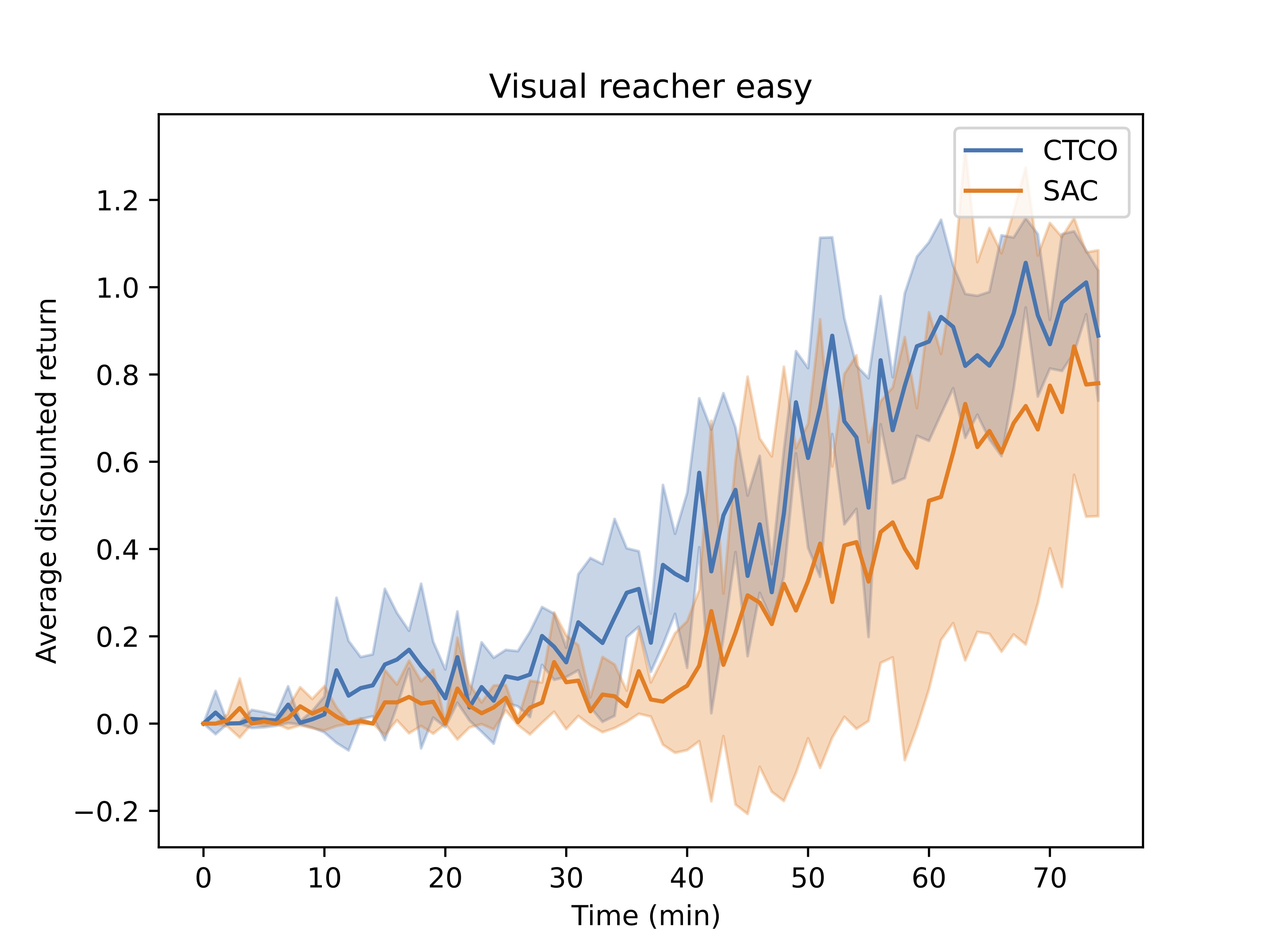}}
    \hfill
    \subfloat[\label{fig:franka_task_result_1}]{
    \includegraphics[width=0.5\textwidth]{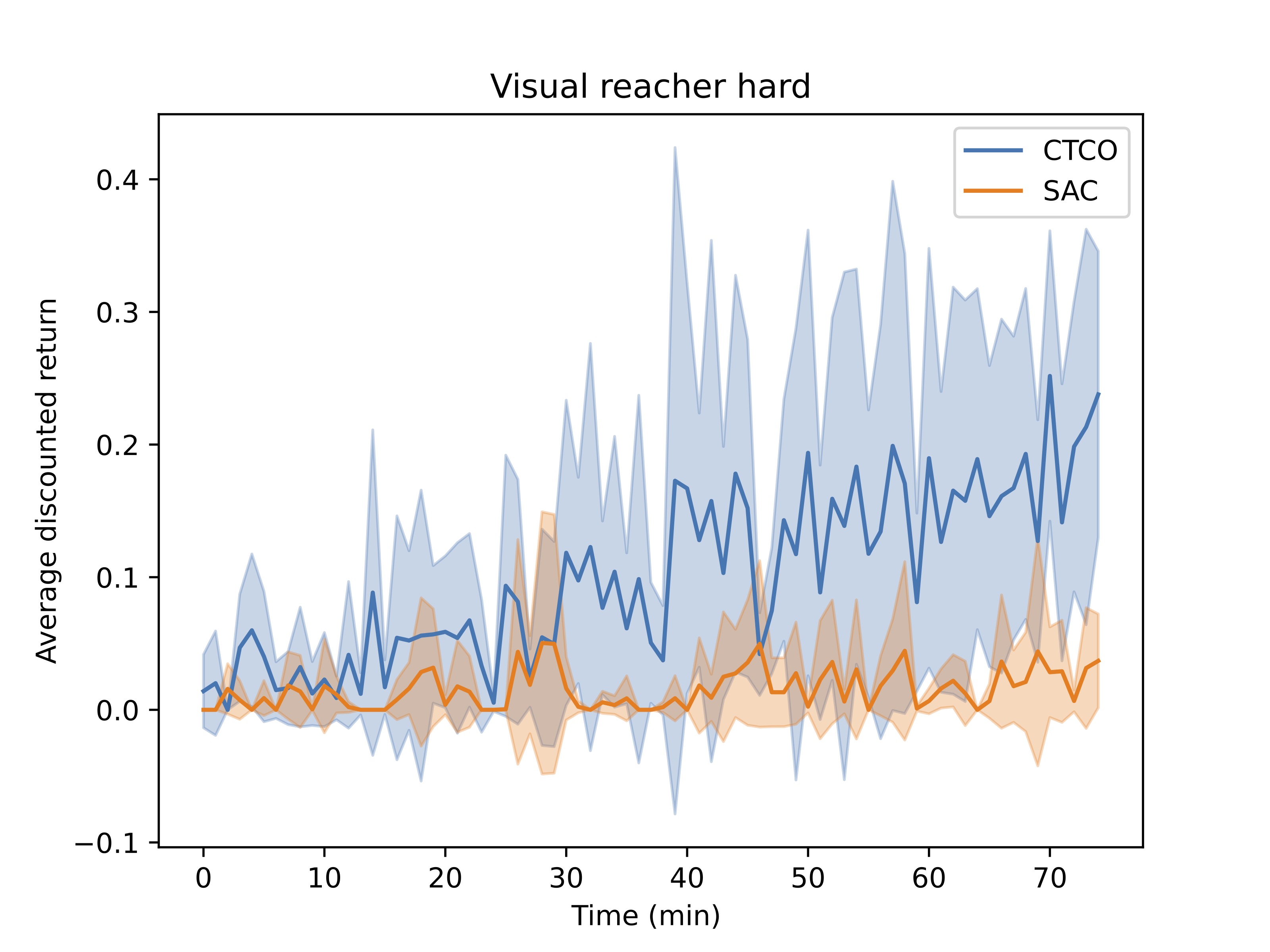}}
    \caption{Demonstration of visual reacher task in (a) and learning performances versus task real time ( reset time is excluded) in (b,c). CTCO reaches higher performances w.r.t SAC in both easy and hard versions of visual reacher. Shaded areas depict 95\% confidence intervals.}
    
\end{figure}

To test the ability of CTCO to work in a real-world scenario, we designed a sparse reward visual-reaching task. Although target reaching can be efficiently solved by using classic robotic techniques such as visual servoing, object detection, and planning, it is still interesting to analyze how an RL agent performs in such tasks. Learning from visual inputs in real-world scenarios without any policy pretraining is in fact very challenging. Visual inputs are subject to noisy observations, and learning algorithms must be highly efficient to learn the task in a reasonable time. 
For this task, we use a $7$ DoF Franka-Emika Panda manipulator shown in Fig.~\ref{fig:franka_task_setup}. We attach an RGB camera to the robot's wrist. A bean bag is randomly placed on the table, and the goal of the robot is to get close enough to the bean bag while it appears in the wrist camera image. The observation consists of an $80 \times 60$ RGB image and joint configurations. The agent controls the robot joint velocities ($7$ dimensions) at $50$Hz. Note that the image observation is sampled at $25$ Hz from the camera and upsampled to $50$ Hz by repetition. Each episode takes 8s to complete. The sparse reward $R(\state,\action)$ is computed as
\begin{align}
    R(\state, \action) = \begin{cases}
    1 & \text{if} \ \rho(\state) \geq 0.015 \\
    0 & \text{otherwise}
    \end{cases} \nonumber 
\end{align}
with \begin{align}
    \rho(\state) & =\frac{1}{w \times h} \sum_{p \space \in \text{red pixels}}(0.5 - |p_x|)(0.5-|p_y|) \nonumber 
\end{align}
 where $\rho(\state)$ is a metric that shows how big and close to the center the bean bag is in the image. $w = 80$, $h= 60$ are image dimensions, and $p_x, p_y \in [-0.5, 0.5]$ are normalized coordinates of red pixels in the camera image.
The robot has the bean bag attached to its wrist by a string. This allows the robot to set the position of the bean bag randomly on the table at the beginning of each episode. The position of the end-effector is bounded to a box of dimensions $30\times 50 \times 30 \ cm^3$. We have two versions of tasks, easy and hard with the difference in the reset. In the easy version, the object is placed uniformly over an area of $20 \times 30 \ cm^2$, but the reset in the hard version is not uniform and places the object unpredictably.

In this experiment, we test SAC against CTCO with $\beta_h = 0.02$, $n_\text{RBF}=2$ for easy and hard visual reacher. Both actor and critics, which are identical for the two algorithms except for the actor's last layer, have a convolutional encoder as the first layer followed by fully connected layers. Interactions and agent updates happen in separate processes to allow real-time control, similarly to \cite{yuan2022asynchronous}. 
Figure~\ref{fig:franka_task_result_2} and ~\ref{fig:franka_task_result_1} show the average and confidence interval of discounted returns over 5 runs. This result indicates that CTCO benefits from the temporal exploration introduced by the continuous options, while SAC fails due to ineffective exploration, especially when the reset is not robust. In the video included in supplementary material, we observe how our algorithm performs the visual reaching using extended smooth actions.

\section{Limitations}
While removing the decision frequency hyper-parameter, our algorithm suffers from two drawbacks: it introduces two new hyper-parameters and has open-loop options. The two new hyper-parameters define the number of RBFs composing the options and the high-frequency penalization. The agent's performance can be sensitive to the choice of these hyperparameters for different reward scales and task complexity levels.
In our work, options have open-loop policies. Therefore, they cannot adapt to unexpected state changes (stochastic environments). For instance, we show in the accompanying video that in the task of visual reaching with a moving object, when running a policy pre-trained with static objects, if the object location is changed too fast CTCO may fail in tracking the object. Technically, the policy can counteract this deficit by choosing low-duration options (thus increasing the frequency of the feedback loop). However, lower-duration policies are undesirable since they complicate learning. This limitation can be compensated by using closed-loop sub-policies or implementing termination policies as in \cite{park_time_2021}.

\section{Conclusion And Future Work}
Most reinforcement learning algorithms are defined in discrete time in which the system is agnostic of what happens between one action and the next. Moreover, the decision of which action to apply is taken at fixed time intervals. When those intervals are too short, the learning agent needs to make many consecutive decisions before reaching its goal, making the problem harder to solve. In contrast, when decisions are not frequent enough, the system can become uncontrollable. In our paper, we propose a reinforcement learning framework where the agent selects new sub-policies as options with variable durations. This approach enhances the algorithm's robustness with respect to the underlying interaction frequency and promotes high-level, smooth exploration. Empirical results underpin this robustness and show the effectiveness of our algorithm in sparse reward settings and for robotic manipulation.
As for future work, we will include a termination mechanism that allows the agent to select a new option before the natural termination of the previous one to counteract unforeseen events due to stochasticity in the environment.
\section{Acknowledgments}
We sincerely acknowledge the support and funding provided by CCIA Chairs, RLAI Lab, Amii, and NSERC of Canada, which made this research possible. We also acknowledge Huawei Noah's Ark Lab for generously donating the Franka Robot arm.

\bibliographystyle{IEEEtran}
\bibliography{zotero}
\newpage
\section{Appendix}
\subsection{Continuous time policy gradient theorem and proof}
Assume parameterizing the policy by $\theta $, According to the bellman equation we have gradient of $Q$ w.r.t to $\theta$ as:

\begin{align*}
    &\nabla_{\theta} Q(\state, \omegavec,  d)=\\
    &\nabla_\theta \underset{\state', \omegavec', d'}{\mathbb{E}}\bigg[
    R(\state, \omegavec, d) - \beta_h + \gamma(d) (Q(\state',\omegavec', d')
    \\
    &\aligntoright{+\beta_E \mathcal{H}(\pi_\theta(.,.|\state')))\bigg]}
    \\&=\nabla_\theta \underset{\state', \omegavec', d'}{\mathbb{E}}\bigg[
    R(\state, \omegavec, d) - \beta_h + \gamma(d) (Q(\state',\omegavec', d') 
    \\&\aligntoright{-\beta_E \log(\pi_\theta(\omegavec',d'|\state')))\bigg]}
    \\&= \underset{\state', \epsilon'}{\mathbb{E}}\bigg[ \gamma(d) \bigg(\nabla_{\omegavec} Q(\state', \omegavec',  d') \nabla_{\theta} f^{\omegavec}_{\theta}(\state', \epsilon')+
    \\&\aligntoright{\nabla_{d} Q(\state', \omegavec',  d') \nabla_{\theta} f^{d}_{\theta}(\state', \epsilon')\bigg) + \gamma(d) \nabla_{\theta} Q(s', \omegavec',d')}
    \\&\aligntoright{-\gamma(d) \nabla_{\theta} \beta_E \log(\pi_\theta(\omegavec',d'|\state')) \bigg]}\tag{By reparameterization trick}.
\end{align*}
We can recursively replace $\nabla_\theta Q(\state', \omegavec', d') \text{ and obtain}$
\begin{align}
    &\nabla_{\theta} Q(\state, \omegavec,  d)=\\ &\underset{\mu_\pi}{\mathbb{E}}\bigg[ \sum_{i=0}^{\infty}  (\prod^{i}_{j=0} \gamma(d_j)) \bigg( \nabla_{\omegavec} Q(\state_{i+1}, \omegavec_{i+1},  d_{i+1}) \nabla_{\theta} f^{\omegavec}_{\theta}(\state_{i+1}, \epsilon_{i+1})  \nonumber
    \\&\aligntoright{+\nabla_{d} Q(\state_{i+1}, \omegavec_{i+1},  d_{i+1}) \nabla_{\theta} f^{d}_{\theta}(\state_{i+1}, \epsilon_{i+1}) }\nonumber
    \\&\aligntoright{-\beta_E \nabla_{\theta} \log(\pi_\theta(\omegavec_{i+1}, d_{i+1}|\state_{i+1}))\bigg)} \nonumber
    \\&\aligntoright{|_{\omegavec_{i+1}=f^{\omegavec}_{\theta}(\state_{i+1}, \epsilon_{i+1}),d_{i+1}=f^{d}_{\theta}(\state_{i+1}, \epsilon_{i+1})} \bigg]} \nonumber
    \\&\text{for}\quad \state_0 = \state, \omegavec_0 = \omegavec , d_0=d.\label{eq:q_recursive}
\end{align}
Hence, 
 
 \begin{theorem}[Continuous-Time Continuous-Option Policy Gradient]
 \label{theorem-Continuous-Time Continuous-Option Policy Gradient}
 	Consider a CT-MDP and a sampling process for $\state_i, \omegavec_i,  d_i$ as described in Section~\ref{sec:method}. The gradient of the objctive w.r.t. to the policy parameter is
 	\begin{align*}
 		&\nabla_\theta J_{\pi} = \underset{\mu_\pi}{\mathbb{E}}\bigg[ \sum_{i=0}^{\infty}\left( \prod_{j=0}^{i-1}\gamma(d_j)\right) \bigg(\nabla_{\omegavec} Q(\state_i, \omegavec_i,  d_i) \nabla_{\theta} f^{\omegavec}_{\theta}(\state_i, \epsilon_i) 
        \\&\aligntoright{+ \nabla_{d} Q(\state_i, \omegavec_i,  d_i) \nabla_{\theta} f^{d}_{\theta}(\state_i, \epsilon_i))-\beta_E \nabla_{\theta} \log(\pi_\theta(\omegavec_i, d_i|\state_i))\bigg)}\\
        &\aligntoright{|_{\omegavec_i=f^{\omegavec}_{\theta}(\state_i, \epsilon_i), d_i=f^{d}_{\theta}(\state_i, \epsilon_i)} \bigg]}
 	\end{align*}
 	where $\gamma(d_i) = e^{-\tau d_{i}}$ (note that $\prod_{j=0}^{-1} \gamma(d_j) = 1$). 
 \end{theorem}
 Since the sampling variables $\omegavec_{i},  d_i, \dots$ are Markov, we can assume that there is a discounted stationary distribution $\zeta^\rho$ from which we can sample them i.i.d. and obtain the same result.
 
Proof: 
\begin{align*}
    J_{\pi} &= \underset{\mu_\pi}{\mathbb{E}}\bigg[ \sum_{i=0}^{\infty}\left( \prod_{j=0}^{i-1}\gamma(d_j)\right) (R(\state_i, \omegavec_i, d_i) - \beta_h 
    \\&\quad\quad\quad\quad\quad\quad\quad\quad\quad\quad\quad\quad\quad\quad+\mathcal{H}(\pi_\theta(.,.|\state_i))\bigg]\\
    &=\underset{\state_0, \omegavec_0, d_0}{\mathbb{E}}\bigg[ Q(\state_0, \omegavec_0, d_0) - \log\pi_\theta(\omegavec_0, d_0|\state_0))\bigg].
\end{align*}
Then
\begin{align*}
    &\nabla_\theta J_{\pi} = \nabla_\theta \underset{\state_0, \omegavec_0, d_0}{\mathbb{E}}\bigg[ Q(\state_0, \omegavec_0, d_0) - \log\pi_\theta(\omegavec_0, d_0|\state_0))  \bigg]
    \\&= \underset{\state_0, \epsilon_0}{\mathbb{E}}\bigg[\nabla_\theta Q(\state_0, \omegavec_0,  d_0)
    + \nabla_{\omegavec} Q(\state_0, \omegavec_0,  d_0)\\
    &\aligntoright{
    \nabla_{\theta} f^{\omegavec}_{\theta}(\state_0, \epsilon_0) 
    +\nabla_{d} Q(\state_0, \omegavec_0,  d_0) \nabla_{\theta} f^{d}_{\theta}(\state_0, \epsilon_0))}
    \\&\quad\quad\quad\quad\quad\quad\quad\quad\quad\quad\quad\quad -\beta_E \nabla_{\theta} \log(\pi_\theta(\omegavec_0, d_0|\state_0)) \bigg]\tag{By reparameterizaiton ${\omegavec_0=f^{\omegavec}_{\theta}(\state_0, \epsilon_0), d_0=f^{d}_{\theta}(\state_0, \epsilon_0)}$}.
\end{align*}
Replacing $\nabla_\theta Q(\state_0, \omegavec_0, d_0) \text{ using equation }$ \ref{eq:q_recursive} results in:
\begin{align*}
    &\nabla_\theta J_{\pi} = \mathbb{E}_{\mu_\pi}\bigg[ \sum_{i=0}^{\infty} \prod_{j=0}^{i-1}\gamma(d_j)\bigg(\nabla_{\omegavec} Q(\state_i, \omegavec_i,  d_i) \nabla_{\theta} f^{\omegavec}_{\theta}(\state_i, \epsilon_i) 
    \\&\quad\quad\quad\quad\quad\quad\quad\quad\quad\quad\quad+\nabla_{d} Q(\state_i, \omegavec_i,  d_i) \nabla_{\theta} f^{d}_{\theta}(\state_i, \epsilon_i))\\
    &\quad\quad-\beta_E \nabla_{\theta} \log(\pi_\theta(\omegavec_i, d_i|\state_i))\bigg)|_{\omegavec_i=f^{\omegavec}_{\theta}(\state_i, \epsilon_i), d_i=f^{d}_{\theta}(\state_i, \epsilon_i)} \bigg].
\end{align*}

Discounting can be dropped to stochastically sample from the discounted distribution, then we have

\begin{align*}
    \nabla_\theta J_{\pi} &= \mathbb{E}_{\mu_\pi}\bigg[  \nabla_{\omegavec} Q(\state, \omegavec,  d) \nabla_{\theta} f^{\omegavec}_{\theta}(\state, \epsilon) 
    \\&\quad\quad\quad\quad\quad\quad\quad\quad\quad\quad+ \nabla_{d} Q(\state, \omegavec,  d) \nabla_{\theta} f^{d}_{\theta}(\state, \epsilon))\\
    &\quad\quad\quad-\beta_E \nabla_{\theta} \log(\pi_\theta(\omegavec, d_i|\state))|_{\omegavec=f^{\omegavec}_{\theta}(\state, \epsilon), d=f^{d}_{\theta}(\state, \epsilon)} \bigg].
\end{align*}\label{sec:appendix}

\end{document}